\RequirePackage{amsmath}
\documentclass[runningheads]{eccv2020kit/llncs}
\usepackage{graphicx}
\usepackage{comment}
\usepackage[colorlinks,breaklinks=true,bookmarks=false]{hyperref}
\hypersetup{linkcolor=black} 

\usepackage{amsmath,amssymb}    
\usepackage{hyphenat}
\usepackage{color}
\usepackage{mwe}
\usepackage{algorithm}          
\usepackage{algpseudocode}      
\usepackage{subfigure}
\usepackage{subfiles}
\usepackage{amssymb}
\usepackage{pifont}
\usepackage[normalem]{ ulem }
\usepackage{soul}
\newcommand{\xmark}{\ding{55}}%
\begin{document}
\pagestyle{headings}
\mainmatter
\def\ECCVSubNumber{5669}  

\title{Unsupervised Domain Adaptation in the Dissimilarity Space for Person Re-identification} %

\titlerunning{UDA in the Dissimilarity Space for Person ReID}
%

\author{Djebril Mekhazni \and
Amran Bhuiyan \and
George Ekladious \and
Eric Granger}
\authorrunning{D. Mekhazni et al.}
%
\institute{LIVIA, Dept. of Systems Engineering \\ \'Ecole de technologie sup\'erieure, Montreal, Canada \\ \email{\{djebril.mekhazni, amran.apece\}@gmail.com,  \{george.ekladious, eric.granger\}@etsmtl.ca}\\}
\maketitle

\begin{abstract}
Person re-identification (ReID) remains a challenging task in many real-word video analytics and surveillance applications, even though state-of-the-art accuracy has improved considerably with the advent of deep learning (DL) models trained on large image datasets. Given the shift in distributions that typically occurs between video data captured from the source and target domains, and absence of labeled data from the target domain, it is difficult to adapt a DL model for accurate recognition of target data. DL models for unsupervised domain adaptation (UDA) are commonly designed in the feature representation space. We argue that for pair-wise matchers that rely on metric learning, e.g., Siamese networks for person ReID, the UDA objective should consist in aligning pair-wise dissimilarity between domains, rather than aligning feature representations. Moreover, dissimilarity representations are more suitable for designing open-set ReID systems, where identities differ in the source and target domains.
In this paper, we propose a novel Dissimilarity-based Maximum Mean Discrepancy (D-MMD) loss for aligning pair-wise distances that can be optimized via gradient descent using relatively small batch sizes. From a person ReID perspective, the evaluation of D-MMD loss is straightforward since the tracklet information (provided by a person tracker) allows to label a distance vector as being either within-class (within-tracklet) or between-class (between-tracklet). This allows approximating the underlying distribution of target pair-wise distances for D-MMD loss optimization, and accordingly align source and target distance distributions. 
Empirical results with three challenging benchmark datasets show that the proposed D-MMD loss decreases as source and domain distributions become more similar. Extensive experimental evaluation also indicates that UDA methods that rely on the D-MMD loss can significantly outperform baseline and state-of-the-art UDA methods for person ReID. The dissimilarity space transformation allows to design reliable pair-wise matchers, without the common requirement for data augmentation and/or complex networks. 
Code is available on GitHub link: \url{https://github.com/djidje/D-MMD}


\end{abstract}

\section{Introduction}


Person re-identification (ReID) refers to the task of determining if a person of interest captured using a camera has the same identity as one of the candidates in the gallery, captured over different non-overlapping camera viewpoints.  It is a key task in object recognition, drawing significant attention due to its wide range of applications, from video surveillance to sport analytics.

Despite the recent advances of ReID with DL models~\cite{TripletLoss,BoT,Auto-ReID,CrossEntropy,st-reID}, and the availability of large amounts of labeled training data, person ReID still remains a challenging task due to the non-rigid structure of the human body, the different perspectives with which a pedestrian can be observed, the variability of capture conditions (e.g., illumination, blur), occlusions and background clutter. In practical video surveillance scenarios, the uncontrolled capture conditions and distributed camera viewpoints can lead to considerable intra-class variation, and to high inter-class similarity. The distribution of image data captured with different cameras and conditions may therefore differ considerably, a problem known in the literature as domain shift~\cite{nguyen2020joint,wang2018deep}. Given this domain shift, state-of-the-art DL models that undergo supervised training with a labeled image dataset (from the source domain) often generalize poorly for images captured in a target operational domain, leading to a decline in ReID accuracy. 

Unsupervised domain adaptation (UDA) seeks resolve the domain shift problem by leveraging unlabeled data from the target domain (e.g., collected during a calibration process), in conjunction with labeled source domain data, to bridge the gap between the different domains. UDA techniques rely on different approaches, ranging from the optimization of a statistical criterion to the integration of an adversarial network, in order to learn robust domain-invariant representations from source and target domain data. Recently, several UDA methods have been proposed for pair-wise similarity matchers, as found in person ReID~\cite{SPGAN,PDA-Net,TJ-AIDL,MSMT17,CameraAwareSimilarity,HHL,ECN}. Common UDA approaches for metric learning employ (1) clustering algorithms for pseudo-labeling of the target data in the feature space, or (2) aligning feature representations of source and target data (either by minimizing some domain discrepancy or adversarial loss)~\cite{wang2018deep}. These feature-based approaches are suitable for closed-set application scenarios, where the source and target domains share the same label space. However, this is not the case in open-set scenarios, where real-world person ReID systems are applied. 

In this paper, we present a new concept for designing UDA methods that are suitable for pair-wise similarity matching in open-set person ReID scenarios. Instead of adapting the source model to unlabeled target samples in the feature representation space, UDA is performed in the dissimilarity representation space. As opposed to the common feature space, where a dimension represents a feature value extracted from one sample (i.e., a vector represents this sample measured over all features), the  dissimilarity space consists of dissimilarity coordinates where each dimension represents the difference between two samples measured for a specific feature (i.e., a vector represents the Euclidean distance between two samples). Accordingly, the multiple clusters that represent different classes (i.e., ReID identities) in the feature representation space, are transformed to only two clusters that represent the pair-wise within- and between-class distances. This transformation is more suitable for open-set ReID problems, when identities differ between the source and target domains, since the new label space has only two labels -- pair-wise similar or dissimilar.  Aligning the pair-wise distance distributions of the source and target domains in the dissimilarity space results in a domain-invariant pair-wise matcher.  

The dissimilarity representation concept was recently introduced in \cite{ekladious2020dual}, where a pseudo-labeling approach was proposed for UDA in still-to-video face recognition. This approach provided descent UDA results for problems with a limited domain shift. As a specific realization of the proposed concept, this paper focuses on a discrepancy-bases approach for  dissimilarity-based UDA, that can provide a high level of accuracy for challenging problems with significant domain shift, as in ReID applications. To this end, we propose a variant of the common Maximum Mean Discrepancy (MMD) loss that is tailored for the dissimilarity representation space. The new Dissimilarity-based MMD (D-MMD) loss exploits the structure of intra- and inter-class distributions to align the source and target data in the dissimilarity space.  It leverages tracklet\footnote{A tracklet correspond to a sequence of bounding boxes that are captured over time for a same person in a camera viewpoint, and obtained using a person tracker.} information to approximate the pair-wise distance distribution of the target domain, and thus estimate a reliable D-MMD loss for alignment of source and target distance distributions.

\begin{figure}[t]
\centering
\includegraphics[width=1\textwidth]{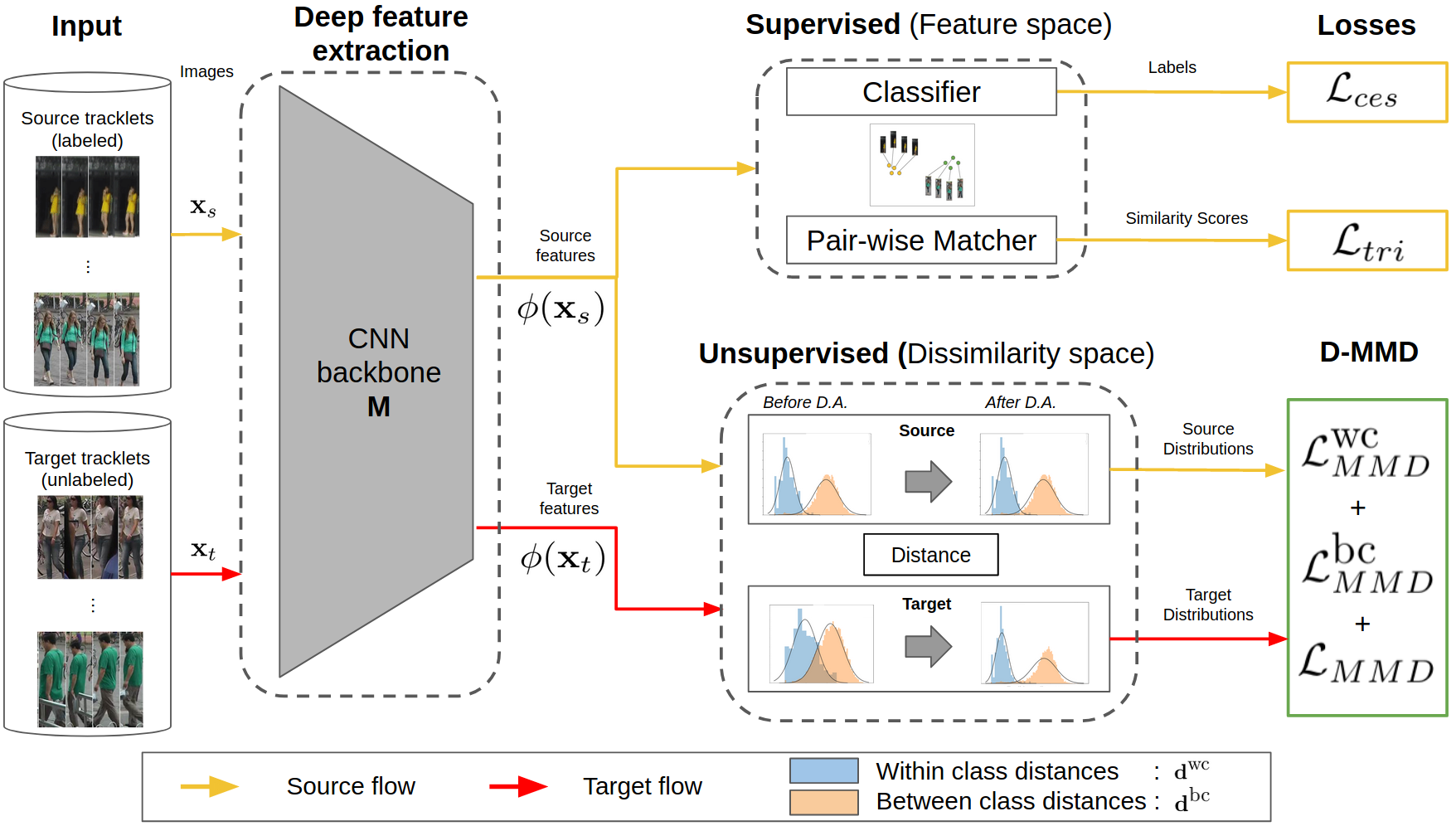}
\caption{\small Deep learning model for UDA using the proposed D-MMD loss. Labeled source images and unlabeled target images are input. First, the DL model for ReID undergoes supervised learning with source images. Upon reaching convergence, the backbone CNN can produce deep features from source and target images. Within-class (WC) and between-class (BC) dissimilarity distributions are produced for source and target domain data. Then, the D-MMD loss is applied between WC (resp. BC) source and WC (resp. BC) target. Supervised losses are also employed to ensure model stability.}
\label{global_framework}
\end{figure}

This paper contributes a novel D-MMD loss for UDA of DL models for person ReID. This loss allows to learn a domain-invariant pair-wise dissimilarity space representation, and thereby bridge the gap between image data from source and target domains (see Fig. \ref{global_framework}). 
An extensive experimental analysis on three benchmark datasets indicates that minimizing the proposed D-MMD loss allows to align the source and target data distributions, which substantially enhances the recognition accuracy across domains. It also allows for designing reliable pair-wise matchers across domains, without the traditional requirement for data augmentation and/or complex networks.

\section{Unsupervised Domain Adaptation for ReID}


UDA focuses on adapting a model such that it can generalize well on an unlabeled target domain data while using a labeled source domain dataset. DL models for UDA seek to learn discriminant and domain-invariant representations from source and target data. They are generally based on either adversarial-, discrepancy-, or reconstruction-based approaches~\cite{wang2018deep}. UDA methods have received limited attention in ReID because of their weak performance on benchmarks datasets compared to their supervised counterparts. Relying on a large-amount of annotated image data, and leveraging the recent success of deep convolutional networks, supervised ReID approaches~\cite{bhuiyan2020pose,chen2019mixed,BoT,Auto-ReID,sun2018beyond} have shown a significant performance improvement, but UDA performance drops drastically when tested on different datasets and large domain shifts. To deal with this issue, representative methods use either clustering-based approach or domain-invariant feature learning based approach. 

In clustering-based approaches~\cite{PUL,PDA-Net}, unlabeled target data are clustered to generate pseudo-labels, and then the network is optimized using the pseudo-labeled target data. Accordingly, performance of these approaches highly depend on the accuracy of clustering algorithms, and low accuracy  can result in the propagation of noisy labels, and a corrupted model. In contrast, the domain-invariant feature learning based approaches
~\cite{bousmalis2016domain,li2017mmd,deepCORAL,tzeng2017adversarial,MMDweight} learn domain-invariant features.  One approach is to define a discrepancy loss function that measures the domain shift in the feature space so that minimization of this loss decreases the domain shift, such in CORAL~\cite{deepCORAL}, MMD GAN~\cite{li2017mmd}, and WMMD~\cite{MMDweight}. Another approach for producing domain-invariant feature representations is through adversarial training, by penalizing a classifier's ability to differentiate between source and target representations~\cite{bousmalis2016domain,tzeng2017adversarial}.  

These approaches either employ pseudo-labeling using a specific set of labels (classes) that exist in the source domain, or represent samples of specific individuals similarly in both source and target domains. Therefore, these  approaches are more suitable for closed-set application scenarios, where the source and target domains share the label space.  Accordingly, these approaches can be ineffective when applied to real-world person ReID applications that generally correspond to an open-set scenario. Indeed, individuals that appear in the target operational domain are typically different than those in design detests, or during the calibration phase.  

To overcome these limitations of domain-invariant feature learning approaches \sloppy, a different category of methods generate synthetic labeled data by transforming the source data to their style representative of target data~\cite{SPGAN,MSMT17,HHL,ECN}. However, performance of these approaches completely depends on the image generation quality. Other methods in the literature use labeled source data to train an initial deep ReID model, and then refine the trained model by clustering the target data~\cite{UTAL,TAUDL,CameraAwareSimilarity}. These methods achieve a lower performance as they do not leverage the labeled source data to guide the adaptation procedure. Moreover, all aforementioned methods ignore the valuable knowledge that can be inferred from the underlying relations among target samples.

This paper addresses the limitations of the existing UDA methods for ReID through transferring the design space from the common feature representation space to the dissimilarity representation space, where open-set models can be easily adapted. This allows aligning the pair-wise distance distributions of the source and target domains.  More specifically, this approach differs from the literature in two main aspects: (1) Unlike \cite{li2017mmd,deepCORAL,MMDweight}, we proposed to use D-MMD loss by exploiting the advantages of intra- and inter-class distributions along with global distributions. This allows dealing with the open-set application scenario exist in person ReID. (2) Our proposed approach does not rely on synthetic data augmentation as in~\cite{SPGAN,MSMT17,HHL,ECN}, nor on the sensitivity of clustering algorithms as in~\cite{PUL,PDA-Net}.

\section{Proposed Method}

In this paper, a novel Dissimilarity-based Maximum Mean Discrepancy (D-MMD) loss is proposed for UDA of ReID systems. Rather  than  aligning  source and target domains feature space, our D-MMD loss allows for the direct alignment of pair-wise distance distributions between domains. This involves j
ointly aligning the pair-wise distances from within-class distributions, as well as distances from between-class distributions. Both of these component contribute to accurate UDA for ReID systems based on a pair-wise similarity matcher, and have not been considered in other state-of-the-art methods.  The proposed  D-MMD loss allows to optimize pair-wise distances through gradient descent using relatively small batches.
 

Fig. \ref{global_framework} shows a DL model for UDA that relies on our D-MMD loss.  For training, 
images $\textbf{x}_s \in \textbf{X}_s$ are sampled from the source domain $\mathcal{D}_s$, while images $\textbf{x}_t \in \textbf{X}_t$ are sampled from the target domain $\mathcal{D}_t$. During UDA, the CNN backbone model $\mathcal{M}$ is adapted to produce a discriminant feature representation $\phi(\textbf{x}_s)$ (resp. $\phi(\textbf{x}_t$)) for input images, and the distances between input feature vectors allows estimating WC or BC distributions.

The underlying relations between source and target domain tracklets are employed to compute distributions of Euclidean distances based on samples of same identity (WC), $\textbf{d}^{\mbox{\small wc}}$, and of  different identities (BC), $\textbf{d}^{\mbox{\small bc}}$. The D-MMD loss $\mathcal{L}_{D-MMD}$ seeks to align the distance distributions of both domains through back propagation.  The overall loss function $\mathcal{L}$ for UDA is:
\begin{equation}
    \mathcal{L} = \mathcal{L}_{\mbox{Supervised}} + \mathcal{L}_{D-MMD}
    \label{Final loss}
\end{equation}
During inference, the ReID system performs pair-wise similarity matching. It is therefore relevant to optimize in the similarity space, and align target similarity distribution with well-separated intra/inter-class distribution from $\mathcal{D}_s$. The rest of this section provides additional details on the $\mathcal{L}_{\mbox{Supervised}}$ and $\mathcal{L}_{D-MMD}$ loss functions.

\subsection{Supervised Loss:}

A model $\mathcal{M}$ is trained through supervised learning on source data $\textbf{X}_s$ using a combination of a softmax cross-entropy loss with label smoothing regularizer ($\mathcal{L}_{\mbox{ces}}$) \cite{CrossEntropy} and triplet loss ($\mathcal{L}_{\mbox{tri}}$) \cite{TripletLoss}. $\mathcal{L}_{\mbox{ces}}$ is defined by Szegedy et al. \cite{CrossEntropy} as:
\begin{equation}
    \mathcal{L}_{\mbox{ces}} = (1 - \epsilon) \cdot \mathcal{L}_{\mbox{ce}} + \frac{\epsilon}{N},
    \label{CrossEntropyFormula}
\end{equation}
where $N$ denotes total number of classes, and $\epsilon$ $\in [0, 1]$ is a hyper-parameter that control the degree of label smoothing. $\mathcal{L}_{\mbox{ce}}$ is defined as:
\begin{equation}
    \mathcal{L}_{\mbox{ce}} =
    \frac{1}{K}\sum_i^K
    -\log \left( \frac{ \exp ( {\textbf{W}^T_{y_i} \textbf{x}_i + b_{y_i}} ) } {\sum_{j=1}^N \exp{ ( \textbf{W}_{j}^T\textbf{x}_i + b_j} ) }   \right)
\end{equation}
where $K$ is the batch size.  Class label $y_{i} \in \{1, 2, ..., N\}$ is associated with training image $\textbf{x}_{i}$, the $i^{th}$ training image. Weight vectors $\textbf{W}_{yi}$ and bias $b_{yi}$ of last fully connected (FC) layer corresponds to class $y$ of $i^{th}$ image. $\textbf{W}_{j}$ and $b_{j}$  are weights and bias of last FC corresponding of the $j^{th}$ class ($j \in [1, N]$). $\textbf{W}_j$ and $\textbf{W}_{y_i}$ are respectively the $j^{th}$ and $y_i^{th}$ column of $\textbf{W} = \{ w_{ij} : i = 1, 2, ..., F; j = 1, 2, ..., N\}$, where $F$ is the size of the last FC layer.

Triplet loss is also employed with hard positive/negative mining as proposed by Hermans et al. \cite{TripletLoss}, where batches are formed by randomly selecting a person, and then sampling a number of images for each person. For each sample, the hardest positive and negative samples  are used to compute the triplet loss:
\begin{equation}\label{BH}
	\begin{aligned}
	\mathcal{L}_{\mbox{tri}}&= \frac{1}{N_s} \sum_{\alpha=1}^{N_s} \big[m + \operatorname*{max} (d \left(\textbf{$\phi$}(\textbf{x}_\alpha^i),\textbf{$\phi$}(\textbf{x}_p^i))\right)- \operatorname*{min}_{i \neq j} (d \left(\textbf{$\phi$}(\textbf{x}_\alpha^i),\textbf{$\phi$}(\textbf{x}_n^j))\right) \big]_{+}
	\end{aligned}
\end{equation} where, $\big[ . \big] _{+} = \max(.,0)$, $m$  denotes a margin, $N_s$ is the set of all hard triplets in the mini-batch, and $d$ is the Euclidean distance. $\textbf{x}_j^i$ corresponds to the $j^{th}$ image of the $i^{th}$ person in a mini-batch. Subscript $\alpha$ indicates an anchor image, while $p$ and $n$ indicate a positive and negative image with respect to that same specific anchor. $\textbf{$\phi$}(\textbf{x})$ is the feature representation of an image $\textbf{x}$.
 


For our supervised loss, we combine both the above  losses.
\begin{equation}
    \mathcal{L}_{\mbox{supervised}} = \mathcal{L}_{\mbox{ces}} + \lambda \cdot  \mathcal{L}_{\mbox{tri}}
    \label{Loss supervised}
\end{equation}
where $\lambda$ is a hyper-parameter that weights the contribution of each loss term.


The softmax cross-entropy loss $\mathcal{L}_{\mbox{ces}}$ defines the learning process as a classification task, where each input image is classified as one of the known identities in the training set. The triplet loss $\mathcal{L}_{\mbox{tri}}$ allows to optimise an embedding where feature vectors are less similar different for inter-class images, and more similar intra-class images.

\subsection{Dissimilarity-based Maximum Mean Discrepancy (D-MMD):}
\label{subsec:Dissimilarity-based Maximum Mean Discrepancy (D-MMD}
After training the model $\mathcal{M}$, we use it to extract feature representations from each source image $\textbf{x}_s \in$  $\textbf{X}_{s}$, $\phi(\textbf{x}_s)$, and target image $\textbf{x}_t \in$  $\textbf{X}_{t}$, $\phi(\textbf{x}_t)$. Then, the within-class distances, e.g., Euclidean or $L_2$ distances, between each different pair of images $\textbf{x}_i^u$ and $\textbf{x}_i^v$ of the same class $i$ are computed:
\begin{equation}
    d^{\mbox{\small wc}}_{i}(\textbf{x}_{i}^u,\textbf{x}_{i}^v)= || \phi(\textbf{x}_{i}^u) 
    - \phi(\textbf{x}_{i}^v) ||_2 , \; u \neq v
    \label{wc-euclidean-distances}
\end{equation}
where $\phi(.)$ is the backbone CNN feature extraction, and $x_{i}^u$ is the image $u$ of the class $i$. Similarly, the between-class distances are computed using each different pair of images $\textbf{x}_i^u$ and $\textbf{x}_j^z$ of the different class $i$ and $j$:
\begin{equation}
    d^{\mbox{\small bc}}_{i,j}(\textbf{x}_{i}^u, \textbf{x}_{j}^z) = || \phi(\textbf{x}_{i}^u) - \phi(\textbf{x}_j^z) ||_2 , \; i \neq j \; \& \; u \neq z
    \label{bc-euclidean-distances}
\end{equation}
Then, $\textbf{d}^{\mbox{\small wc}}$ and $\textbf{d}^{\mbox{\small bc}}$ are defined as the distributions of all distance values $d^{\mbox{\small wc}}_{i}$ and $d^{\mbox{\small bc}}_{i,j}$, respectively, in the dissimilarity space.

The within-class (WC) and between-class (BC) distance samples of the source domain are computed directly using the source labels, so they capture the exact pair-wise distance distribution of the source domain. On the other hand, given the unlabeled target data, we leverage the tracklet information provided by a visual object tracker. We consider the frames within same tracklet as WC samples, and frames from different tracklets as BC samples. It is important to note that such tracklet information provide us with an approximation of the pair-wise distance distribution of the target domain since it lacks intra-class pairs from  different tracklets or cameras. 


Maximum Mean Discrepancy (MMD) \cite{MMD} metric is used to compute the distance between two distribution:
\begin{multline}
        MMD(P(A), Q(B)) = \frac{1}{n^2}\sum_{i=1}^n\sum_{j=1}^n k(a_i, a_j)
        \\ + \frac{1}{m^2}\sum_{i=1}^m\sum_{j=1}^m k(b_i, b_j) 
        - \frac{2}{nm}\sum_{i=1}^n\sum_{j=1}^m k(a_i, b_j)
        \label{MMD}
\end{multline}
where $A$ (resp. $B$) is the source (resp. target) domain and $P(A)$ (resp. $Q(B)$) is the distribution of the source (resp. target) domain. $k(.,.)$ is a kernel (e.g. Gaussian) and $a_i$ (resp. $b_i$) is sample $i$ from $A$ (resp. $B$). $n$ and $m$ are number of training examples from $P(A)$ and $Q(B)$, respectively. 

\begin{figure}[ht]
\label{fig:Distributions}
\centering
\hspace*{\fill}
    \subfigure [Source distributions]
    {
        \includegraphics[width=0.3\linewidth]{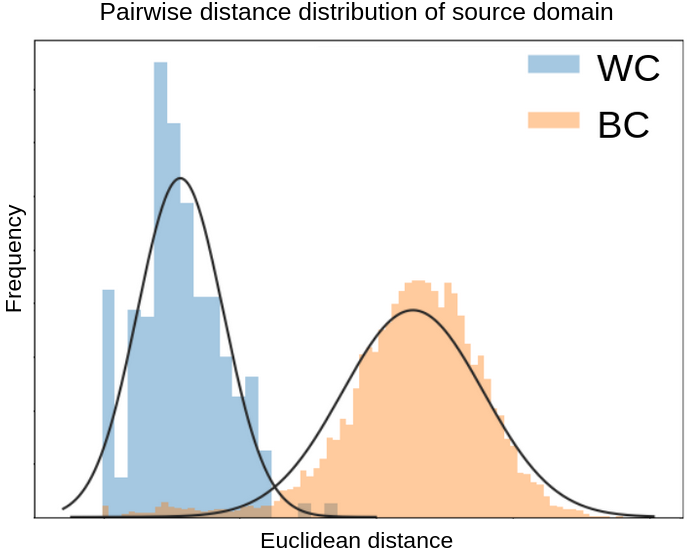}
        \label{fig:Distri-source}
    }
    \subfigure [Target without DA]
    {
        \includegraphics[width=.3\linewidth]{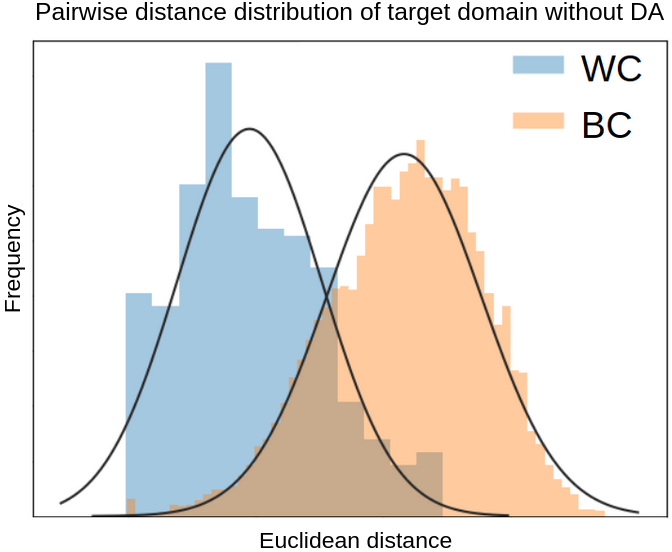} 
        \label{fig:Distri-without-da}
    }
    \subfigure [Target with D-MMD]
    {
        \includegraphics[width=.3\linewidth]{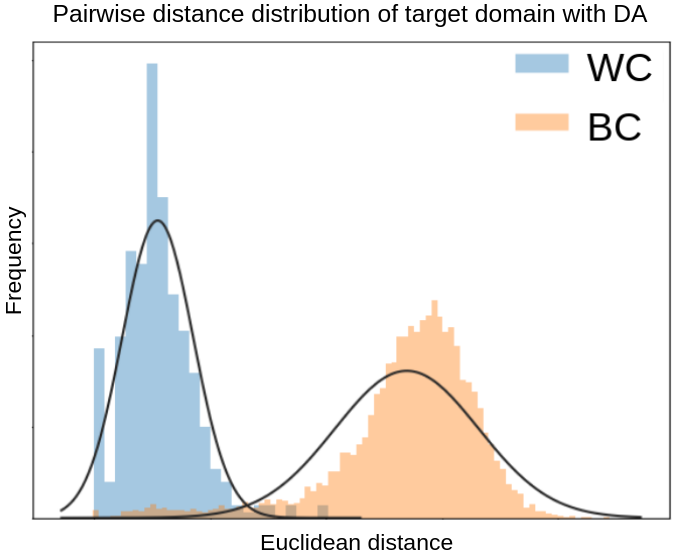}
        \label{fig:Distri-with-da}
    }
    
    \small{\caption{Fig. \ref{fig:Distri-source} shows that the dissimilarity representation of the WC (blue) and BC (orange) distributions, where BC has larger Euclidean distances than WC because the model produces features closer for samples from same identities  than that for samples from different identities. Fig. \ref{fig:Distri-without-da} shows a significant overlap when target data  are represented using the initial source model, due to the intrinsic domain shift. Figure \ref{fig:Distri-with-da} shows that the target BC and WC distributions become aligned with the source distributions (Fig. \ref{fig:Distri-source}) after performing UDA.}}
   
   
\end{figure}

To evaluate the divergence between two domains, MMD metric is applied to measure the difference from features produced by the source and target models are different using:
\begin{equation}
    \mathcal{L}_{MMD} = MMD(\mathcal{S}, \mathcal{T}) 
    \label{MMD global}
\end{equation}
$\mathcal{S}$ ($\mathcal{T}$) is defined as the distribution of the sources (target) images  $\textbf{X}_{s}$ ($\textbf{X}_{t}$) represented in the feature space. 

Our method relies on the application of the MMD in the dissimilarity representation space instead of the common feature representation space.
%
%
We define the $\mathcal{L}_{MMD}^{\mbox{\small wc}}$  and $\mathcal{L}_{MMD}^{\mbox{\small bc}}$ loss terms as follows: 
\begin{equation}
    \mathcal{L}_{MMD}^{\mbox{\small wc}} = MMD(\textbf{d}^{\mbox{\small wc}}_s, \textbf{d}^{\mbox{\small wc}}_t)
    \label{MMD WC}
\end{equation}
\begin{equation}
    \mathcal{L}_{MMD}^{\mbox{bc}} = MMD(\textbf{d}^{\mbox{\small bc}}_s, \textbf{d}^{\mbox{\small bc}}_t)
    \label{MMD BC}
\end{equation}
Minimizing the above terms aligns the pair-wise distance distributions of the source and target domains, so that pair-wise distances from different domains are not deferential, and hence the source model works well in the target domain. Finally, our unsupervised loss function can be expressed as: 
\begin{equation}
    \mathcal{L}_{D-MMD} = \mathcal{L}_{MMD}^{\mbox{\small wc}}  + \mathcal{L}_{MMD}^{\mbox{\small bc}} + \mathcal{L}_{MMD}
    \label{MMD loss}
\end{equation}

\small{\begin{algorithm}[htbp]
    \begin{algorithmic}
        \caption{UDA training strategy based on the D-MMD loss.}
        \label{algorithm: D-MMD algorithm}
        
        \Require labeled source data $\textbf{X}_s$, and unlabeled target data $\textbf{X}_t$
        
        \State \textbf{Load} source data $\textbf{X}_s$, and \textbf{Initialize} backbone model $\mathcal{M}$
        \For{$l \in [1, N^s]$ epochs}
            \For{each mini-batch $B_s \subset \textbf{X}_s$}
                \State \textbf{1)} \textbf{Compute} $\mathcal{L}_{\mbox{ces}}$ with Eq. \ref{CrossEntropyFormula}, and $\mathcal{L}_{\mbox{tri}}$ with Eq. \ref{BH}
                \State \textbf{2)} \textbf{Optimize} $\mathcal{M}$ 
            \EndFor
        \EndFor

        \State \textbf{Load} target data $\textbf{X}_t$, and \textbf{Load} backbone model $\mathcal{M}$

         \For{$l \in [1, N^u]$ epochs}
        
            \For{each mini-batch $B_s \subset \textbf{X}_s$ each mini-batch $B_t \subset \textbf{X}_t$}
            
                \State \textbf{1)} \textbf{Generate} $\textbf{d}^{\mbox{\small wc}}_s$ with Eq. \ref{wc-euclidean-distances}, $\textbf{d}^{\mbox{\small bc}}_s$ with Eq. \ref{bc-euclidean-distances} and $B_s$
                \State \textbf{2)} \textbf{Generate} $\textbf{d}^{\mbox{\small wc}}_t$ with Eq. \ref{wc-euclidean-distances}, $\textbf{d}^{\mbox{\small bc}}_t$ with Eq. \ref{bc-euclidean-distances} and $B_t$
                \State \textbf{3)} \textbf{Compute} $\mathcal{L}_{D-MMD}(B_s, B_t)$ with Eq. \ref{MMD loss}
                \State \textbf{4)} \textbf{Compute} $\mathcal{L}_{\mbox{Supervised}}$ with Eq. \ref{Loss supervised} using $B_s$
                \State \textbf{5)} \textbf{Optimize} $\mathcal{M}$ based on overall $\mathcal{L}$ (Eq. \ref{Final loss})
                
            \EndFor
            
        \EndFor
    \end{algorithmic}
\end{algorithm}}

Algorithm \ref{algorithm: D-MMD algorithm} presents a UDA training strategy based on the D-MMD loss. Firstly, a supervised training phase runs for $N^s$ epochs and produces a reference model $\mathcal{M}$ using the source domain data. Then, an unsupervised training phase runs for $N^u$ epochs and aligns the target  and source pair-wise distance distributions by minimizing the D-MMD loss terms defined by Eq. \ref{MMD global}, Eq. \ref{MMD WC}, and Eq. \ref{MMD BC}.
Note that the supervised loss $\mathcal{L}_{Supervised}$ is evaluated during domain adaptation to ensure the model $\mathcal{M}$ remains aligned to a reliable source distribution $\mathcal{S}$ over training iterations remaining.


Evaluating the D-MMD loss during the UDA training strategy involves computing the distances among each pairs of images. The computational complexity can be estimated as the number of within-class and between-class distance calculations. Assuming a common batch size of $|B|$ for training with source and target images and a number of occurrence of the same identity $N_o$, the total number of distance calculations is: 
\begin{equation}
    N_{\mbox{distances}}   = N_{\mbox{distances}}^{wc}        +  N_{\mbox{distances}}^{bc} 
                    = (N_o-1)! \frac{|B|}{N_o}  +  N_o (\frac{|B|}{N_o}-1 )^2
    \label{Number operation WC}
\end{equation}

\section{Results and Discussion}

\subsection{Experimental methodology:}

For the experimental validation, we employ three challenging person ReID \sloppy datasets, Market-1501 \cite{Market1501}, DukeMTMC \cite{DukeMTMC} and MSMT17 \cite{MSMT17}, and compare our proposed approach with state-of-the-art generative (GAN), tracklet-based, and domain adaptation methods for unsupervised person ReID.

\small{\begin{table}
\caption{Properties of the three challenging datasets used in our experiments. They are listed according to their  complexity (number of images, persons, cameras, and capture conditions, e.g., occlusions and illumination changes).}
\begin{tabular}{|c||c|c|c|c|c|c|c|c|}
\hline
\textbf{Datasets} 
& \begin{tabular}[c]{@{}c@{}}\#  \\ IDs\end{tabular} 
& \begin{tabular}[c]{@{}c@{}}\#  \\ cameras\end{tabular} 
& \begin{tabular}[c]{@{}c@{}}\#  \\ images\end{tabular} & \begin{tabular}[c]{@{}c@{}}\# train \\ (IDs)\end{tabular} & \begin{tabular}[c]{@{}c@{}}\# gallery \\ (IDs)\end{tabular} & \begin{tabular}[c]{@{}c@{}}\# query \\ (IDs)\end{tabular} & \begin{tabular}[c]{@{}c@{}}Annotation \\ method\end{tabular} & \begin{tabular}[c]{@{}c@{}}Crop \\ size\end{tabular} \\ \hline \hline
\begin{tabular}[c]{@{}c@{}}Market-\\ 1501\end{tabular}   & 1501          & 6          & 32217     & \begin{tabular}[c]{@{}c@{}}12936\\ (751)\end{tabular}     & \begin{tabular}[c]{@{}c@{}}15913\\ (751)\end{tabular}       & \begin{tabular}[c]{@{}c@{}}3368\\ (751)\end{tabular}      & \begin{tabular}[c]{@{}c@{}}semi-automated\\ (DPM)\end{tabular}     & 128x64                                               \\ \hline
\begin{tabular}[c]{@{}c@{}}Duke-\\ MTMC\end{tabular}     & 1812          & 8          & 36441     & \begin{tabular}[c]{@{}c@{}}16522\\ (702)\end{tabular}     & \begin{tabular}[c]{@{}c@{}}17661\\ (1110)\end{tabular}      & \begin{tabular}[c]{@{}c@{}}2228\\ (702)\end{tabular}      & manual                                                    & variable                                                 \\ \hline
MSMT17                                                   & 4101          & 15         & 126441    & \begin{tabular}[c]{@{}c@{}}32621 \\ (1041)\end{tabular}    & \begin{tabular}[c]{@{}c@{}}82161 \\ (3060)\end{tabular}     & \begin{tabular}[c]{@{}c@{}}11659 \\ (3060)\end{tabular}   & \begin{tabular}[c]{@{}c@{}}semi-automated\\ (Faster R-CNN)\end{tabular}   & variable                                                 \\ \hline
\end{tabular}
\label{Datasets}
\end{table}}

Table. \ref{Datasets} describes three datasets for our experimental evaluation -- Market1501, DukeMTMC and MSMT17. the \textbf{Market-1501} \cite{Market1501} dataset comprises  labels generated using Discriminatively Trained Part-Based Models (DPM) \cite{DPM}. It provides a realistic benchmark, using 6 different cameras, and around ten times more images than previously published datasets.  The \textbf{DukeMTMC} \cite{DukeMTMC} dataset is comprised of videos captured outdoor at the Duke University campus from 8 cameras. \textbf{MSMT17} \cite{MSMT17} is the largest and most challenging ReID dataset. It is comprised of indoor and outdoor scenarios, in the morning, noon and afternoon, and each video is captured over a long period of time.  

For the supervised training, a Resnet50 architecture model is pretrained on ImageNet until convergence for both Hard-Batch Triplet and Softmax Cross-Entropy loss functions. Source domain videos are utilized for supervised training and evaluation. Then, the source and target training videos are used to perform UDA of the source model. Features are extracted  from  images of both domains using the Resnet50 CNN backbone (with a 2048 features vector size). To compute the BC and WC distributions, we randomly selected 4 occurrences of each class within  batches of size 128. Given the nature of data, the tracklets are subject to greater diversity, with images from different viewpoints. The $D-MMD$ is then computed as described in Section 3, and backpropagation is performed using an Adam optimizer with a single step scheduler, decreasing the learning rate by 10 (initially 0.003) after every 20 epochs. In all steps, every image is resized to $256 \times 128 $ before being processed.

Table \ref{Supervised Results} reports the upper bound accuracy for our datasets. To obtain this reference, we leveraged labeled source and target image data for supervised training. The ResNet50 model is initially trained using data from a first person ReID dataset (source domain), and then it is fine-tuned with training data from a second ReID datasets (target domain). Accuracy is computed with the target test sets of respective ReID datasets. We employed cross entropy loss with label smoothing regularizer \ref{CrossEntropyFormula} with  $\epsilon = 0.1$, and triplet loss with a margin  $m = 0.3$. To train DukeMTMC and Market1501, 30 epochs are required, while MSMT17 requires 59 epochs due to its larger-scale and complexity. 
Results in Table~\ref{Supervised Results} confirms  that MSMT17 is the most challenging  dataset and shows  lowest performance (63.2 \% rank-1 accuracy). 



Instead of optimizing the  number of occurrences (frames) in a tracklet as a hyper-parameter, we had to use a fixed number (4 occurences) since  the  experimental  datasets can sometime include only this number of frames per tracklet, and also for fair comparison with the SOA results. The metrics  used for performance evaluation  are the mean average precision (mAP), and  rank-1, rank-5, rank-10 accuracy from the Cumulative Match Curve (CMC). 

\subsection{Ablation study:}

Table. \ref{Ablation-study} shows the impact on accuracy of the different loss terms. It is clear that $\mathcal{L}_{MMD}^{\mbox{bc}}$ and $\mathcal{L}_{MMD}^{\mbox{wc}}$ provide important information, with a slight improvement for the between-class (BC) component. Moreover, results show that a combination of both losses produces better results than when each term is employed separately. Moreover, while the classic feature-based $\mathcal{L}_{MMD}$ had insignificant impact when employed separately (as observed in \cite{PDA-Net}), it helps when combined with the other terms. This can be explained by the fact that $\mathcal{L}_{MMD}$ suffers from ambiguous association while dealing with domain shift that exists in open-set scenarios. Nevertheless, when the domain gap decreases to a reasonable limit (with the help of the proposed dissimilarity-based loss terms $\mathcal{L}_{MMD}^{\mbox{bc}}$ and $\mathcal{L}_{MMD}^{\mbox{wc}}$), the feature-based loss starts to contributing ReID accuracy. 

\begin{table}[t]
\centering
\caption{Upper bound accuracy obtained after training on source data, and then fine-tuning on target data.  Accuracy is measured with target training data.}
\begin{tabular}{|l||c|c|c|c|}
\hline
\multicolumn{1}{|c||}{\textbf{Dataset}}       & \multicolumn{4}{c|}{\textbf{Accuracy}} \\ \cline{2-5} 
\multicolumn{1}{|c||}{source $\longrightarrow$ target} & rank-1   & rank-5   & rank-10  & mAP   \\ \hline \hline
DukeMTMC $\longrightarrow$ Market1501            & 89.5     & 95.6      & 97.1  & 75.1  \\ \hline
Market1501 $\longrightarrow$ DukeMTMC            & 79.3     & 89.3      & 92.0  & 62.7  \\ \hline
DukeMTMC $\longrightarrow$ MSMT17                & 63.2     & 77.5      & 82.0  & 33.9  \\ \hline
\end{tabular}
\label{Supervised Results}
\end{table}

\small{\begin{table}[b]

\caption{Ablation Study. Impact on accuracy of individual loss terms when transferring between the DukeMTMC and Market1501 domains.  (The lower bound accuracy refers is obtained with the ResNet50 model trained on source data, and tested on target data, without domain adaptation.)}
\centering
\begin{tabular}{|c|c|c|c|c|c|c|c|c|}
\hline
\textbf{Setting} & \multicolumn{4}{c|}{\textbf{Loss Functions}} & \multicolumn{2}{l|}{\begin{tabular}[l]{@{}l@{}} Source: Duke\\ Target: Market\end{tabular}} & \multicolumn{2}{l|}{\begin{tabular}[l]{@{}l@{}}Source: Market\\ Target:  Duke\end{tabular}} \\ \cline{2-9} 
& $\mathcal{L}_{sup}$    & $\mathcal{L}_{MMD}^{\mbox{wc}}$  & $\mathcal{L}_{MMD}^{\mbox{bc}}$  & $\mathcal{L}_{MMD}$  & rank-1  & mAP & rank-1 & mAP \\ \hline \hline
Lower Bound      & \checkmark     & \xmark      & \xmark      & \xmark      &                  36.1                        &                  16.1                       &              23.7                            &                     12.3                    \\ \hline
A                       & \checkmark     & \checkmark     & \xmark      & \xmark      &             45.8                             &                      24.3                   &                30.3                          &                    16.6                     \\ \hline
B                        & \checkmark     & \xmark      & \checkmark      & \xmark     &                51.8                          &                   28.4                      &                 45.6                         &                29.2                         \\ \hline
C                        & \checkmark    & \checkmark     & \checkmark      & \xmark     &                 66.6                         &                45.4                         &          60.5                                &                  42.9                       \\ \hline
D                        & \checkmark    & \checkmark     & \checkmark      & \checkmark     &                 \textbf{70.6}                         &                \textbf{48.8}                         &          \textbf{63.5}                                &                  \textbf{46.0}                       \\ \hline

\end{tabular}

\label{Ablation-study}
\end{table}}




From source DukeMTMC to target Market1501,  the margin of improvement while considering only the WC component over the baseline are 9.7\% for Rank-1 accuracy and 8.2\% for mAP and for only BC component 15.7 \% for Rank-1 accuracy and 12.1 \% for mAP. From source Market1501 to target DukeMTMC, we reach for the WC component 6.6 \% for Rank-1 accuracy and 4.3\% for mAP  improvement compared to the baseline when for BC component we obtain 21.9\% for Rank-1 accuracy and 16.9 \% for mAP more than the baseline.

Table. \ref{Ablation-study} shows that a model adapted using only BC information is capable to produce better representation and leads to better results (51.8\% Rank-1 accuracy) than when using only WC (45.8\% Rank-1 accuracy) for the DukeMTMC to Market1501 transfer problem. In general, combining the different terms provides better results than when individual losses are employed. 



\begin{figure}[ht!]
    \centering
    \subfigure[]
    {
        \includegraphics[width=0.3\linewidth]{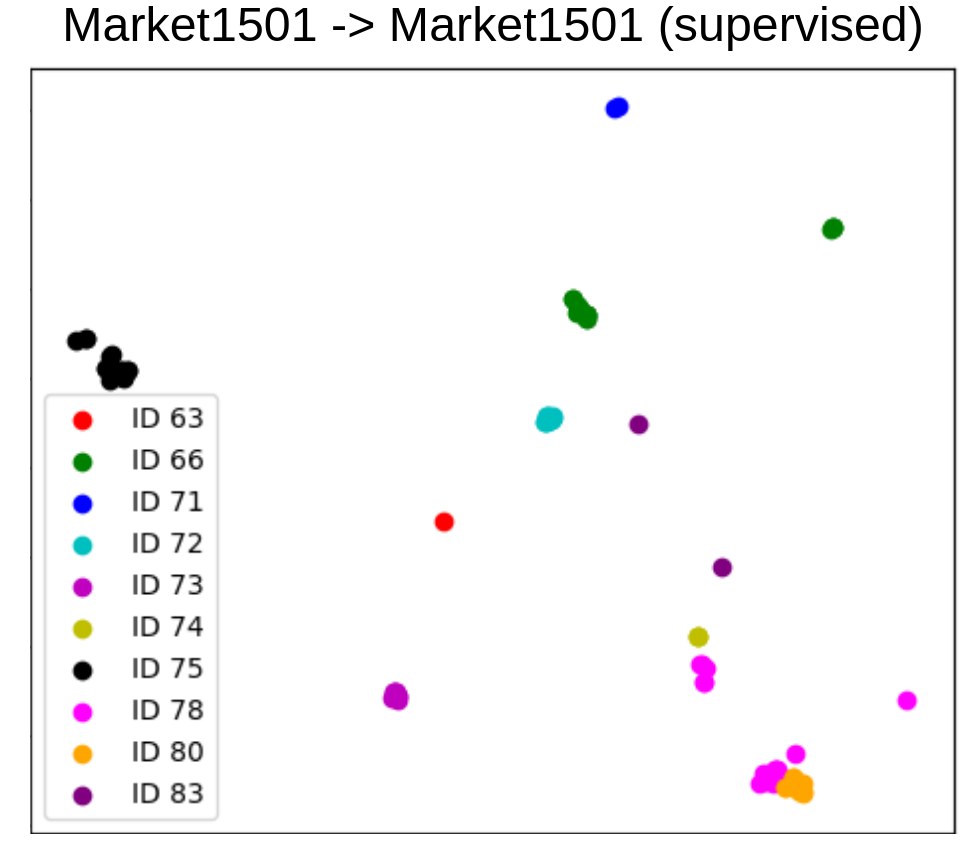}
        \label{fig:first_sub}
    }
    \subfigure[]
    {
        \includegraphics[width=.305\linewidth]{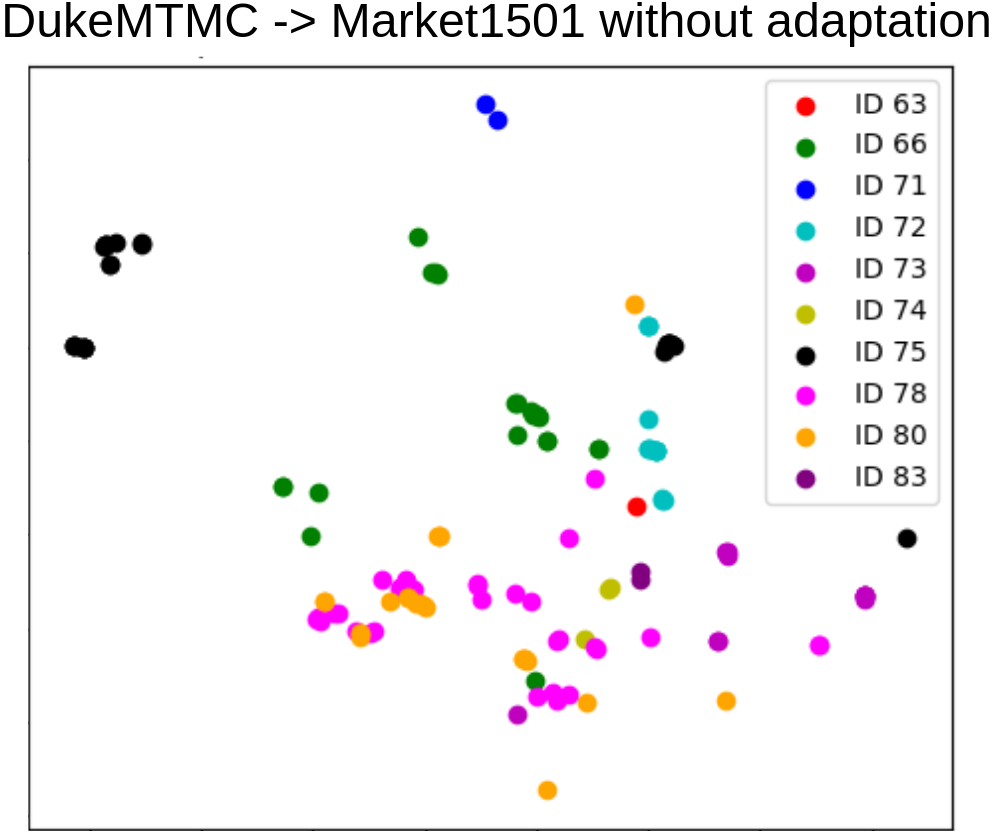}
        \label{fig:second_sub}
    }
    \subfigure[]
    {
        \includegraphics[width=.3\linewidth]{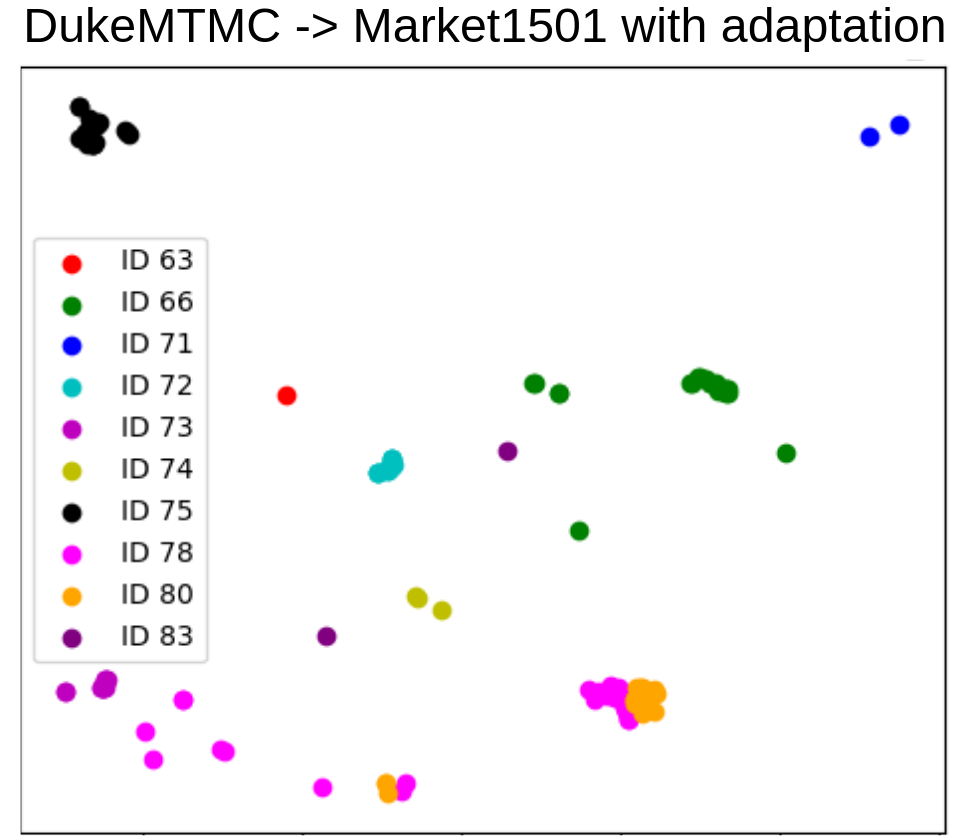}
        \label{fig:third_sub}
    }
    
    \caption{T-SNE visualisations that show the impact of the original domain shift (\ref{fig:first_sub} versus  \ref{fig:second_sub}). Then \ref{fig:second_sub} and \ref{fig:third_sub} show the impact of employing our method to decrease domain shift, and accordingly improving the representation.}
    
    \label{fig:sample_subfigures}
\end{figure}

In Section \ref{subsec:Dissimilarity-based Maximum Mean Discrepancy (D-MMD} (Fig.2),  it is shown clearly that there is large overlap between the intra- and inter-class pair-wise distance distributions when using the initial source representation in the target domain. When applying the proposed method, the overlap significantly decreases and aligned with the source distributions.  Fig \ref{fig:sample_subfigures} shows the reflection of such improvement of the pair-wise distance representation on the actual Re-ID problem. Before applying DA, there is much confusion between person representations which can be improved significantly with applying the proposed method.

\subsection{Comparison with state-of-art methods:}

\small{\begin{table}[htbp]
\centering

\caption{ReID accuracy of the proposed and SOTA methods for UDA using Market1501 as source, and DukeMTMC and MSMT17 as targets. Accuracy is obtained on target datasets.}
\begin{tabular}{l||cccc||cccc||c}
\hline
\multicolumn{1}{c||}{\textbf{Methods}}                                                       & \multicolumn{8}{c||}{\textbf{Source: Market1501}}                                                                               & \multicolumn{1}{c}{\textbf{Conference}} \\ \cline{2-9}
\multicolumn{1}{c||}{}                                                                               & \multicolumn{4}{c||}{\textbf{DukeMTMC}}   & \multicolumn{4}{c||}{\textbf{MSMT17}}  & \multicolumn{1}{c}{\textbf{or Journal}}                            \\  
                                                                                               & r-1           & r-5           & r-10          & mAP           & r-1           & r-5           & r-10          & mAP           &                                                  \\ \hline \hline
Lower Bound                                                                                          & 23.7          & 38.8          & 44.7          & 12.3          & 6.1           & 12.0          & 15.6          & 2.0           &  --                                                 \\
\begin{tabular}[c]{@{}c@{}}PUL\cite{PUL}\end{tabular}                                                                                          & 30.0          & 43.4          & 48.5          & 16.4          & -           & -          & -          & -           & TOMM’18                                                 \\
CFSM \cite{CFSM}                                                                                         & 49.8         & -         & -          & 27.3          & -           & -          & -          & -           & AAAI’19                                                 \\
BUC \cite{BUC}                                                                                          & 47.4          & 62.6          & 68.4          & 27.5          & -           & -          & -          & -           & AAAI’19                                                 \\
ARN \cite{ARN}                                                                                          & 60.2          & 73.9          & 79.5         & 33.5          & -           & -          & -          & -           & CVPR’18-WS                                                 \\
UCDA-CCE \cite{UCDA-CCE}                                                                                          & 47.7          & -         & -          & 31.0         & -           & -          & -          & -          & ICCV’19                                                 \\
PTGAN \cite{MSMT17}                                                                                          & 27.4          & -         & 50.7          & -         & 10.2           & -          & 24.4          & 2.9          & CVPR’18                                                 \\
\begin{tabular}[c]{@{}c@{}}SPGAN+LMP\cite{SPGAN}\end{tabular}                                            & 46.4          & 62.3          & 68.0          & 26.2          & -             & -             & -             & -             & CVPR’18                                          \\
\begin{tabular}[c]{@{}c@{}}HHL\cite{HHL}\end{tabular}                                                  & 46.9          & 61.0          & 66.7          & 27.2          & -             & -             & -             & -             & ECCV’18                                          \\
\begin{tabular}[c]{@{}c@{}}TAUDL\cite{TAUDL}\end{tabular}                                           & 61.7          & -             & -             & 43.5          & 28.4          & -             & -             & 12.5          & ECCV’18                                          \\
\begin{tabular}[c]{@{}c@{}}UTAL\cite{UTAL}\end{tabular}                                            & 62.3          & -             & -             & 44.6          & 31.4          & -             & -             & 13.1          & TPAMI’19                                         \\
\begin{tabular}[c]{@{}c@{}}TJ-AIDL\cite{TJ-AIDL}\end{tabular}                                             & 44.3          & 59.6          & 65.0          & 23.0          & -             & -             & -             & -             & CVPR'18                                       \\
\begin{tabular}[c]{@{}c@{}}Wu et al.\cite{CameraAwareSimilarity}\end{tabular} & 51.5          & 66.7          & 71.7          & 30.5          & -             & -             & -             & -             & ICCV’19                                          \\
\begin{tabular}[c]{@{}c@{}}ECN\cite{ECN}\end{tabular}                                                 & 63.3          & 75.8          & 80.4          & 40.4          & 25.3          & 36.3          & 42.1          & 8.5           & CVPR’19                                          \\
\begin{tabular}[c]{@{}c@{}}PDA-Net\cite{PDA-Net}\end{tabular}                                             & 63.2          & 77.0          & 82.5          & 45.1          & -             & -             & -             & -             & IEEE'19                                          \\
\textbf{D-MMD (Ours)}                                                                                                 & \textbf{63.5} & \textbf{78.8} & \textbf{83.9} & \textbf{46.0} & \textbf{29.1} & \textbf{46.3} & \textbf{54.1} & \textbf{13.5} & -- \\ \hline                                        
\end{tabular}
\label{SOA Source Market1501}
\end{table}}

\small{\begin{table}[htbp]
\centering
\caption{ReID accuracy of the proposed and SOTA methods for UDA using DukeMTMC as source, and Market1501 and MSMT17 as targets. Accuracy is obtained on target datasets.}
\begin{tabular}{l||cccc||cccc||c}
\hline
\multicolumn{1}{c||}{\textbf{Methods}}                                                       & \multicolumn{8}{c||}{\textbf{Source: DukeMTMC}}                                                                                 & \multicolumn{1}{c}{\textbf{Conference}} \\ \cline{2-9}
\multicolumn{1}{c||}{}                                                                               & \multicolumn{4}{c||}{\textbf{Market1501}}                      & \multicolumn{4}{c||}{\textbf{MSMT17}}                          & \multicolumn{1}{c}{\textbf{or Journal}}                            \\  
  & r-1  & r-5  & r-10  & mAP  & r-1 & r-5  & r-10  & mAP   &   \\ \hline \hline
Lower Bound                                                                                          & 36.6          & 54.5          & 62.9          & 16.1          & 11.3          & 20.6          & 25.7          & 3.7           & --                                                \\
\begin{tabular}[c]{@{}c@{}}PUL\cite{PUL}\end{tabular}                                                                                          & 45.5          & 60.7          & 66.7          & 20.5          & -           & -          & -          & -           & TOMM’18                                                 \\
CFSM \cite{CFSM}                                                                                         & 61.2         & -         & -          & 28.3          & -           & -          & -          & -           & AAAI’19                                                 \\
BUC \cite{BUC}                                                                                          & 66.2          & 79.6          & 84.5          & 38.3          & -           & -          & -          & -           & AAAI’19                                                 \\
ARN \cite{ARN}                                                                                          & 70.3         & 80.4          & 86.3         & 39.4          & -           & -          & -          & -           & CVPR’18-WS                                                 \\
UCDA-CCE \cite{UCDA-CCE}                                                                                          & 60.4          & -         & -          & 30.9         & -           & -          & -          & -          & ICCV’19                                                 \\
PTGAN \cite{MSMT17}                                                                                          & 38.6          & -         & 66.1          & -         & 10.2           & -          & 24.4          & 2.9          & CVPR’18                                                 \\
\begin{tabular}[c]{@{}c@{}}SPGAN+LMP \cite{SPGAN}\end{tabular}                                            & 57.7          & 75.8          & 82.4          & 26.7          & -             & -             & -             & -             & CVPR’18                                          \\
\begin{tabular}[c]{@{}c@{}}HHL\cite{HHL}\end{tabular}                                                  & 62.2          &               &               & 31.4          & -             & -             & -             & -             & ECCV’18                                          \\
\begin{tabular}[c]{@{}c@{}}TAUDL\cite{TAUDL}\end{tabular}                                           & 63.7          & -             & -             & 41.2          & 28.4          & -             & -             & 12.5          & ECCV’18                                          \\
\begin{tabular}[c]{@{}c@{}}UTAL\cite{UTAL}\end{tabular}                                            & 69.2          & -             & -             & 46.2          & 31.4          & -             & -             & 13.1          & TPAMI’19                                         \\
\begin{tabular}[c]{@{}c@{}}TJ-AIDL\cite{TJ-AIDL}\end{tabular}                                             & 58.2          & 74.8          & 81.1          & 26.5          & -             & -             & -             & -             & CVPR'18                                       \\
\begin{tabular}[c]{@{}c@{}}Wu et al.\cite{CameraAwareSimilarity}\end{tabular} & 64.7          & 80.2          & 85.6          & 35.6          & -             & -             & -             & -             & ICCV’19                                          \\
\begin{tabular}[c]{@{}c@{}}ECN\cite{ECN}\end{tabular}                                                 & \textbf{75.6} & \textbf{87.5} & \textbf{91.6} & 43.0          & 30.2          & 41.5          & 46.8          & 10.2          & CVPR’19                                          \\
\begin{tabular}[c]{@{}c@{}}PDA-Net\cite{PDA-Net}\end{tabular}                                             & 75.2          & 86.3          & 90.2          & 47.6          & -             & -             & -             & -             & IEEE'19                                          \\
\textbf{D-MMD (Ours)}                                                                                                 & 70.6          & 87.0          & 91.5          & \textbf{48.8} & \textbf{34.4} & \textbf{51.1} & \textbf{58.5} & \textbf{15.3} & -- \\ \hline                                       
\end{tabular}
\label{SOA Source DukeMTMC}
\end{table}}

We compare our approach with state-of-the-art (SOTA) unsupervised  methods on Market-1501, DukeMTMC-reID and MSMT17. Lower Bound refers to the domain shift without any adaptation. Table. \ref{SOA Source Market1501} reports the comparison when tested on DukeMTMC and MSMT17 with Market1501 as the source, and Table. \ref{SOA Source Market1501} reports results when DukeMTMC is the source. 

PUL \cite{PUL} and BUC \cite{BUC} are clustering methods for pseudo-labeling of target data. Such approaches lead to poor performance. We outperform them by a large margin, 16.4\% Rank-1 accuracy and 18.5\% mAP more from Market1501 to DukeMTMC than BUC approach \cite{BUC}. TAUDL \cite{TAUDL} and UTAL \cite{UTAL} are two tracklet-based approaches for unsupervised person ReID.  Due to their fully unsupervised behavior, they obtain worse results than our approach.

We also compare with other UDA approaches: PTGAN \cite{MSMT17}, SPGAN \cite{SPGAN}, ARN \cite{ARN}, TJ-AIDL \cite{TJ-AIDL} (attribute-based), \cite{CFSM} HHL \cite{HHL}, ECN \cite{ECN}, PDA-Net \cite{PDA-Net}, Wu et al. \cite{CameraAwareSimilarity}, UDCA-CCE \cite{UCDA-CCE} (Camera-aware). Most of them are using data augmentation methods \cite{ECN,HHL,MSMT17,SPGAN}. We are not using such techniques which are computationally expensive and require more memory. This  also helps with problems that involve transferring from a small dataset  to a larger and more complex dataset, which reflects a natural real-world application scenario.  For the  DukeMTMC to Market1501 transfer problem, we  notice that DukeMTMC is a better initialization for simpler domains such  Market1501, and it is easier to perform well in that sense (similar phenomenon for all other methods). 

ECN and PDA-Net obtain better results on CMC metrics Rank-1 for this  transfer problem DukeMTMC-Market1501 than ours (ECN \cite{ECN} has 5.0\% more and PDA-net \cite{PDA-Net} has 4.6\% higher Rank-1 accuracy for only 0.5\% and 0.1\% on Rank-5 and rank-10 accuracy regarding ECN). We outperform all other methods in mAP metrics (5.4\% more than ECN).  Since the D-MMD objective is to learn domain-invariant pair-wise dissimilarity representations, so success can be better measured using more global metrics, e.g., mAP, and this is validated by our results, where the proposed method produces best results using this metric. In contrast, CMC top-1 accuracy could be improved by training a pattern classifier (eg. MLP) to process the resulting distance vector. Nevertheless, when considering the opposite transfer problem, i.e., Market1501 to DukeMTMC, which is much more complex, our method provides best results for all metrics (0.3\% on Rank-1, 3\% on Rank-5, 3.5\% on Rank-10 accuracy and 5.6\% on mAP). We also outperform state-of-the-art methods on the most challenging dataset MSMT17 by 4.2\% Rank-1, 9.6\% Rank-5, 11.7\% Rank-10 and 5.1\% mAP when considering source DukeMTMC dataset. Similar results are observed using  Market1501 as source.

\small{\begin{table}[h]
\centering
\caption{UDA accuracy of the proposed versus lower and upper bound approaches when transferring from  MSMT17 (source) to Market1501 and DukeMTMC (targets).}
\begin{tabular}{|l||cccc||cccc|}
\hline
\multicolumn{1}{|c||}{\textbf{Methods}}  & \multicolumn{8}{|c|}{\textbf{Source: MSMT17}} \\ \cline{2-9}
\multicolumn{1}{|c||}{}                           & \multicolumn{4}{c||}{\textbf{  Market1501}} & \multicolumn{4}{c|}{\textbf{  DukeMTMC}} \\ 
                                          & rank-1      & rank-5      & rank-10     & mAP     & rank-1           & rank-5           & rank-10          & mAP                                                       \\ \hline
Lower Bound                                    & 43.2     & 61.4     & 68.6     & 20.7    & 47.4          & 63.7          & 69.2          & 27.5                                                   \\
\textbf{D-MMD (Ours)}                                           & 72.8     & 88.1     & 92.3     & 50.8    & 68.8 & 82.6 & 87.1 & 51.6 \\ 
Upper Bound                                  & 89.5     & 95.6     & 97.1     & 75.1    & 79.3          & 89.3          & 92.0          & 62.7                                                      \\
\hline
\end{tabular}
\label{SOA Source MSMT17}
\end{table}}


The proposed method can provide best performance for problems where the source domain consists in challenging data with high intra-class variability and high inter-class similarity (e.g., MSMT17) as compared to easier target domains (e.g. Market1501 and DukeMTMC). Such transfer problem is less explored in the literature, so in Table. \ref{SOA Source MSMT17} we compare our results with only the lower and upper bounds. With this setup (i.e. source is the most challenging dataset MSMT17) we obtained best results (better than these reported on Tables \ref{SOA Source Market1501} and \ref{SOA Source DukeMTMC}): 77.8\% Rank-1 accuracy and 50.8\% mAP for Market1501 and 68.8\% Rank-1 accuracy and 51.6\% mAP for  DukeMTMC.



\section{Conclusion}

In this paper, we proposed a novel dissimilarity-based UDA approach for person ReID using MMD  loss to reduce the gap between domains in the dissimilarity space. The core idea is to exploit the advantages of using within and between-class distances that effectively capture the underlying relations  between  domains  which  has  never  been  explored  in  the state-of-the-art.  To that end, we align the  within- and between-class distance distributions for the source and target domains to produce effective Re-ID models for the target domain. Experiments on three challenging ReID datasets prove the effectiveness of this new approach as it  outperforms state-of-the-art methods. Moreover, our proposed loss is general and can be applied to  different feature extractors and applications.

%
%

\clearpage

\bibliographystyle{eccv2020kit/splncs04}
\bibliography{eccv2020submission}
\end{document}